\documentclass{nseJournal}

\begin{document}

\title{Mechanistic Interpretability of LoRA-Adapted Language Models for Nuclear Reactor Safety Applications}

\addAuthor{\correspondingAuthor{Yoon Pyo Lee}}{a}

\correspondingEmail{lukeyounpyo@hanyang.ac.kr}

\addAffiliation{a}{Department of Nuclear Engineering, Hanyang University, Seoul, Republic of Korea}

\addKeyword{Explainable AI}
\addKeyword{Large Language Models}
\addKeyword{Nuclear Safety}
\addKeyword{Low Rank Adaptation}
\addKeyword{Neuron Silencing}

\titlePage

\begin{abstract}
The integration of Large Language Models (LLMs) into safety-critical domains, such as nuclear engineering, necessitates a deep understanding of their internal reasoning processes. This paper presents a novel methodology for interpreting how an LLM encodes and utilizes domain-specific knowledge, using a Boiling Water Reactor system as a case study. We adapted a general-purpose LLM (Gemma-3-1b-it) to the nuclear domain using a parameter-efficient fine-tuning technique known as Low-Rank Adaptation. By comparing the neuron activation patterns of the base model to those of the fine-tuned model, we identified a sparse set of neurons whose behavior was significantly altered during the adaptation process. To probe the causal role of these specialized neurons, we employed a neuron silencing technique. Our results demonstrate that while silencing most of these specialized neurons individually did not produce a statistically significant effect, deactivating the entire group collectively led to a statistically significant degradation in task performance. Qualitative analysis further revealed that silencing these neurons impaired the model's ability to generate detailed, contextually accurate technical information. This paper provides a concrete methodology for enhancing the transparency of an opaque black-box model, allowing domain expertise to be traced to verifiable neural circuits. This offers a pathway towards achieving nuclear-grade artificial intelligence (AI) assurance, addressing the verification and validation challenges mandated by nuclear regulatory frameworks (e.g., 10 CFR 50 Appendix B), which have limited AI deployment in safety-critical nuclear operations.
\end{abstract}

\section{INTRODUCTION}
\label{sec:introduction}

The nuclear industry operates under a stringent regulatory and safety culture in which every component and decision-making tool must be transparent, auditable, and verifiable \cite{iaea_insag4}. Foundational principles, such as defense-in-depth, which requires multiple independent layers of protection, and As Low As Reasonably Achievable, which mandates minimizing all risks \cite{nrc_alara}, demand a high level of assurance for any new technology introduced into plant operations. Recent advances in Large Language Models (LLMs), built upon the Transformer architecture \cite{vaswani2017attention}, offer the potential for knowledge management and operational support \cite{brown2020language, touvron2023llama}. However, their deployment in safety-critical environments faces a fundamental barrier: the black-box nature of their reasoning processes, which is in conflict with the industry's core safety principles.

The deployment of software in safety-critical systems within a nuclear power plant must adhere to stringent quality assurance (QA) requirements, which is fundamentally different from its use in non-critical applications. In the United States (U.S.), regulations such as 10 CFR 50, Appendix B, mandate a documented design basis, comprehensive testing, and full traceability of a system's logic and behavior \cite{cfr_10_50_b}. This regulatory framework presents a significant challenge for advanced artificial intelligence (AI) and LLMs. The inherent black-box nature of these models makes it difficult, if not impossible, to verify the internal reasoning path an LLM took to arrive at a recommendation, conflicting with QA standards. This has largely limited AI deployment in nuclear facilities for non-safety roles. Therefore, a new paradigm focusing on internal transparency is required to bridge the gap between the capabilities of modern AI and the safety standards of the nuclear industry. Accordingly, any plant-facing AI must provide traceable, auditable links from external requirements and source documents to its internal computations and final outputs.

The integration of advanced digital systems into nuclear operations is an ongoing effort in the industry. Recent work ranges from the modernization of instrumentation and control systems \cite{naser2003modernizing} to the application of LLMs for transient identification \cite{qi2024multimodal} and simulator assistance \cite{lee2025net}. However, these applications do not resolve the inherent black-box nature of the models. In parallel, foundational safety principles have remained a constant focus, with studies addressing human factors considerations for AI applications in nuclear plants \cite{mohon2024human, nureg_0711}, software quality assurance for nuclear computational tools \cite{williamson2021bison}, and the need for robust fault diagnosis frameworks \cite{qi2023fault}. These studies established the importance of verification and validation (V\&V) processes for safety-critical nuclear software systems, underscoring the need for interpretable AI methods in nuclear applications.

This paper addresses this gap by proposing a methodology based on mechanistic interpretability to enhance the trustworthiness of LLMs in nuclear applications. The goal is to increase the transparency of these opaque models, making their internal reasoning processes examinable and validatable, a prerequisite for qualification in systems important to safety, as outlined in key industrial standards such as the Institute of Electrical and Electronics Engineers (IEEE) Std 7-4.3.2 \cite{ieee_std_7_4_3_2}, and their associated regulatory guidelines, such as the U.S. Nuclear Regulatory Commission (NRC) Regulatory Guide 1.152 \cite{nrc_rg_1152}. Recent initiatives like the NRC's Technology-Inclusive Regulatory Framework \cite{nrc_tirf} recognize the potential benefits of advanced digital systems while emphasizing the need for robust safety assurance methodologies. We demonstrate this approach by adapting a general-purpose LLM \cite{gemma_team2025gemma3} to the specific domain of Boiling Water Reactors (BWRs) using documentation from authoritative bodies like the International Atomic Energy Agency (IAEA) and Organisation for Economic Co-operation and Development (OECD)/Nuclear Energy Agency (NEA) \cite{lam2005bwr, nea2022samg}.

To perform this domain adaptation efficiently without the prohibitive cost of full-model retraining, we leverage Parameter-Efficient Fine-Tuning (PEFT) methods. Although various techniques exist, such as Prefix-Tuning \cite{li2021prefix} and Prompt-Tuning \cite{lester2021power}, our study employed Low-Rank Adaptation (LoRA) \cite{hu2021lora}, a prominent technique known for its effectiveness in specialized tasks \cite{liu2022fewshot}. However, verification remains challenging, even after successful adaptation. The trust gap in AI, where models may exhibit unreliable behavior or hidden biases \cite{zhao2021calibrate}, must be closed before such tools can be used in nuclear operations.

Therefore, our research moves beyond simple fine-tuning to investigate the internal mechanisms of the adapted model. We were guided by the following objectives:
\begin{enumerate}
    \item Identify specific neurons whose activation patterns are systematically altered by LoRA fine-tuning on a corpus of BWR technical data.
    \item Use causal interventions to test the functional importance of these identified neurons, with performance quantitatively measured using the established Bilingual Evaluation Understudy (BLEU) \cite{papineni2002bleu} metrics.
    \item Propose a verifiable, step-by-step methodology for tracing abstract knowledge to the physical substrate within a neural network, thereby enhancing its interpretability.
\end{enumerate}

By demonstrating that domain expertise can be traced to a small, verifiable set of neural components, this study provides a pathway towards achieving nuclear-grade AI assurance. This mechanistic insight is a foundational step towards building the auditable and robust AI systems required for the future of nuclear science and engineering. The remainder of this paper is organized as follows. Section \ref{sec:background} reviews the key AI concepts relevant to this study. Section \ref{sec:methodology} describes our experimental approach, including dataset construction, model adaptation, and our causal analysis techniques. Section \ref{sec:results} presents the findings of the neuron silencing experiments. Section \ref{sec:discussion} interprets these results and discusses their implications for AI in terms of nuclear safety. Finally, Section \ref{sec:conclusion} summarizes our work and suggests directions for future research.

\section{BACKGROUND: AI, INTERPRETABILITY, AND NUCLEAR SAFETY}
\label{sec:background}

This section provides an overview of the key concepts underlying this study. First, it presents the central challenge of using AI in safety-critical domains, and then introduces the foundational AI techniques and interpretability methods employed to address this challenge.

\subsection{Domain Adaptation of Large Language Models}
\label{ssec:domain_adaptation}
An LLM is an AI model, built upon the Transformer architecture, that is pre-trained on vast text datasets to acquire general language and world knowledge \cite{vaswani2017attention, devlin2019pretraining}. To adapt such a generalist model to a specialized field, a process called fine-tuning is used, where the model is further trained on a smaller, domain-specific dataset—in this case, a corpus of BWR technical documents.

However, fine-tuning all of the billions of parameters of the model is not only computationally prohibitive but also risks inducing "catastrophic forgetting," where its valuable pre-trained knowledge is overwritten by new, domain-specific data. We employ LoRA, a prominent PEFT technique that freezes the original model weights and injects small trainable adaptation modules (low-rank matrices) into the Transformer layers \cite{hu2021lora}. This allows for efficient domain adaptation while preserving the model's valuable pre-trained knowledge.

\subsection{Mechanistic Interpretability for Nuclear-Grade Assurance}
\label{ssec:mech_interp}
To meet nuclear-grade assurance requirements, we must move beyond "post-hoc" Explainable AI (XAI) techniques like Local Interpretable Model-agnostic Explanations (LIME) or SHapley Additive Explanations (SHAP), which explain predictions by correlating inputs and outputs without revealing the model's internal workings \cite{ribeiro2016lime, lundberg2017shap}. In contrast, "mechanistic interpretability," the approach adopted in this study, aims to directly analyze and causally verify the internal computational structures— neural circuits—that generate a model's behavior.

In high-reliability fields such as nuclear engineering, verifying how a conclusion was reached is as important as the conclusion itself. Neuronal activation was used as a direct signal for this purpose. A neuron is a fundamental computational unit in the model, and its activation is a numerical value representing its output in response to an input. By tracking which neurons' activations are systematically altered during domain adaptation, we can begin to trace abstract knowledge to a physical, verifiable substrate within the model, a prerequisite for satisfying the stringent V\&V demands of nuclear regulators.

\section{METHODOLOGY}
\label{sec:methodology}

Our investigation into the internal mechanisms of domain adaptation followed a three-stage experimental framework: (1) construction of a domain-specific dataset, (2) PEFT of the base model, and (3) neuron-level analysis through activation tracking and causal intervention.

\subsection{Dataset Construction}
\label{ssec:dataset}
We constructed a bespoke Question-Answering dataset to train and evaluate the model in a realistic nuclear engineering context. The source material for this dataset was a collection of authoritative technical documents covering BWR systems, safety procedures, and simulator specifications from various international bodies. Specifically, we utilized documentation from the OECD NEA on severe accident mitigation \cite{nea2022samg}, an IAEA report on BWR simulators \cite{lam2005bwr}, and system descriptions from the U.S. NRC Technical Training Center \cite{nrc_bwr_systems}, all chosen for their comprehensive and trusted coverage of the BWR domain.

The Question-Answering pairs were generated using a custom Python script that leveraged OpenAI's gpt-4o-2024-08-06 application programming interface (API) \cite{openai_gpt4o_blog}. To ensure high quality and traceability, this process involved several key steps: First, the source texts were segmented into manageable chunks (approx. 800 characters). Second, structured prompts instructed the model to create relevant questions and their corresponding answers based exclusively on the text provided, a constraint designed to prevent factual hallucination. Finally, the model returned the output in a structured JSON format, embedding metadata like the source file for each pair to ensure full traceability. This process yielded a total of 1,304 Question-Answering pairs: 506 pairs for the SAMG documentation, 698 pairs from the simulator reports, and 100 pairs from the systems descriptions.

The resulting dataset was divided using stratified sampling based on the source document to ensure a balanced representation. This approach was critical because each source document covers a distinct sub-domain of BWR knowledge. Stratified sampling ensures these topics are proportionally represented in both the training and evaluation sets, preventing topical bias and enabling a more robust assessment of the model’s overall domain expertise. The dataset was split into training ($80\%$) and evaluation ($20\%$) sets, maintaining the stratified distribution across all source documents.

\subsection{Model Fine-Tuning}
\label{ssec:finetuning}

The base model for our study was Gemma-3-1b-it, a general-purpose LLM \cite{gemma_team2025gemma3}. For domain adaptation, we fine-tuned this model on the training split of our custom Question-Answering dataset using the LoRA \cite{hu2021lora} technique. We selected LoRA among PEFT methods because it freezes the base weights and trains only low-rank adapter matrices, which helps preserve pre-trained knowledge and reduces catastrophic forgetting while enabling efficient adaptation. This locality is concentrated in a small number of adapter parameters, making activation shifts easier to trace and analyze, and allowing targeted causal tests. The LoRA configuration was applied to a comprehensive set of target modules within the model's attention and feed-forward layers: q proj, k proj, v proj, o proj, gate proj, up proj, and down proj. The complete set of hyperparameters used in this process is listed in Table \ref{tab:hyperparams}. This process resulted in the specialized model, referred to as "LoRA" in our experiments.

% Hyperparameter Table
\begin{table}[tbp]
\caption{Fine-tuning Hyperparameters}
\label{tab:hyperparams}
\centering
\begin{tabular}{ll}
\hline
\textbf{Parameter} & \textbf{Value} \\
\hline
\multicolumn{2}{l}{\textit{Model \& Quantization}} \\
Base Model           & Gemma-3-1b-it \\
Compute Data Type    & bfloat16 \\
Precision Type       & FP16 (for training) \\
\hline
\multicolumn{2}{l}{\textit{LoRA Configuration}} \\
Rank (\texttt{r})    & 8 \\
Alpha (\texttt{alpha}) & 16 \\
Dropout              & 0.05 \\
Target Modules       & q proj, k proj, v proj, o proj, \\
                     & gate proj, up proj, down proj \\
\hline
\multicolumn{2}{l}{\textit{Training Arguments}} \\
Optimizer            & AdamW (PyTorch) \\
Learning Rate        & 2e-5 \\
Batch Size (per device) & 1 \\
Gradient Accum. Steps & 8 \\
Effective Batch Size & 8 \\
Epochs               & 2 \\
Warmup Steps         & 50 \\
\hline
\end{tabular}
\end{table}

\subsection{Neuron-Level Analysis}
\label{ssec:analysis}
The core of our analysis involved identifying and probing the neurons responsible for the model's newly acquired domain expertise. To facilitate the extraction of internal activation and the implementation of model interventions, we utilized PyTorch's hook mechanism for fine-grained control over neuron activation. This approach aligns with the principles of mechanistic interpretability, which allows us to probe specific neural circuits.

\subsubsection{Identifying Key Neurons}
The first step in our analysis was to identify the specific neurons responsible for encoding newly acquired domain knowledge. We a priori focus on the final Multilayer Perceptron (MLP) layer, which is the last nonlinear feature-synthesis block that writes into the residual stream immediately before readout. This proximity to the output makes it the most plausible location to detect adaptation-induced features with a favorable signal-to-noise ratio while avoiding a broad multiple-comparison search. We measured the average activation of each neuron in the final Multilayer Perceptron (MLP) layer for both the base Gemma model and our fine-tuned LoRA model across the entire evaluation dataset, and computed the average activation ($\Delta_{\text{act}} = \text{AvgAct}_{\text{LoRA}} - \text{AvgAct}_{\text{Base}}$).

To select a compact core set for causal analysis, we applied an elbow-style criterion to the sorted magnitudes $|\Delta_{\text{act}}|$ within this layer. After ranking all neurons by $|\Delta_{\text{act}}|$, we observed a clear change in slope around the sixth item. Operationally, we retained the smallest $k$ for which the marginal gain fell below $20\%-25\%$ of the first step. We therefore focused on the top six neurons as the most substantially altered components, comprising five amplified ($\Delta_{\text{act}} > 0$) and one suppressed ($\Delta_{\text{act}} < 0$) unit to probe both amplification and suppression effects.

\subsubsection{Causal Intervention via Neuron Silencing}
To test the functional role of these key neurons, we employed a causal intervention technique known as neuron silencing.
Silencing was implemented by attaching a forward hook to the final MLP layer of the model using PyTorch's register forward hook function.
This hook intercepts the layer's output tensor during the forward pass of the model and sets the activation values at the specified neuron indices to zero before they are passed to subsequent layers.
This method allows for targeted, real-time intervention without altering the underlying weights of the model.
For each key neuron identified, we created a modified version of the LoRA model by implementing this silencing hook.
Additionally, we created a model where all identified key neurons were silenced simultaneously ("LoRA-Silenced-Key6").
The effects of these interventions were evaluated using the BLEU metric.

\subsubsection{Statistical Analysis}
To determine whether the observed differences in the performance metric (BLEU) between the model pairs were statistically significant, we employed the Wilcoxon signed-rank test \cite{wilcoxon1945}. This non-parametric test is suitable for comparing paired samples, such as the lists of scores generated by two different models on the same set of questions. A p-value of less than 0.05 was considered the threshold for statistical significance.

\subsection{Performance Evaluation Metric}
\label{ssec:perf_eval_metric}
To quantitatively assess the impact of our causal interventions, we selected the BLEU score as the primary metric. In the safety-critical context of nuclear operations, linguistic precision is not merely a matter of style but a direct component of safety. An answer that is factually correct but phrased ambiguously or imprecisely can lead to operator misinterpretation and subsequent errors. The BLEU score, by calculating n-gram precision against a reference answer, directly measures the textual fidelity and fluency of the model’s output. This makes it a suitable proxy for the clarity and reliability of the information provided, which is a paramount concern in this domain. We compute sentence-level BLEU-4 for each question in our evaluation set and report the mean of these scores. BLEU does not capture factual accuracy, so we interpret it as a clarity and fluency proxy that complements our mechanistic and qualitative analyses.

\section{RESULTS}
\label{sec:results}

This section presents the experimental results. It begins with the identification of neurons affected by fine-tuning, followed by the quantitative outcomes of the neuron-silencing interventions.

\subsection{LoRA Fine-Tuning Induces Neuron Specialization}
\begin{figure}[tbp]
    \centering
    \includegraphics[width=\textwidth]{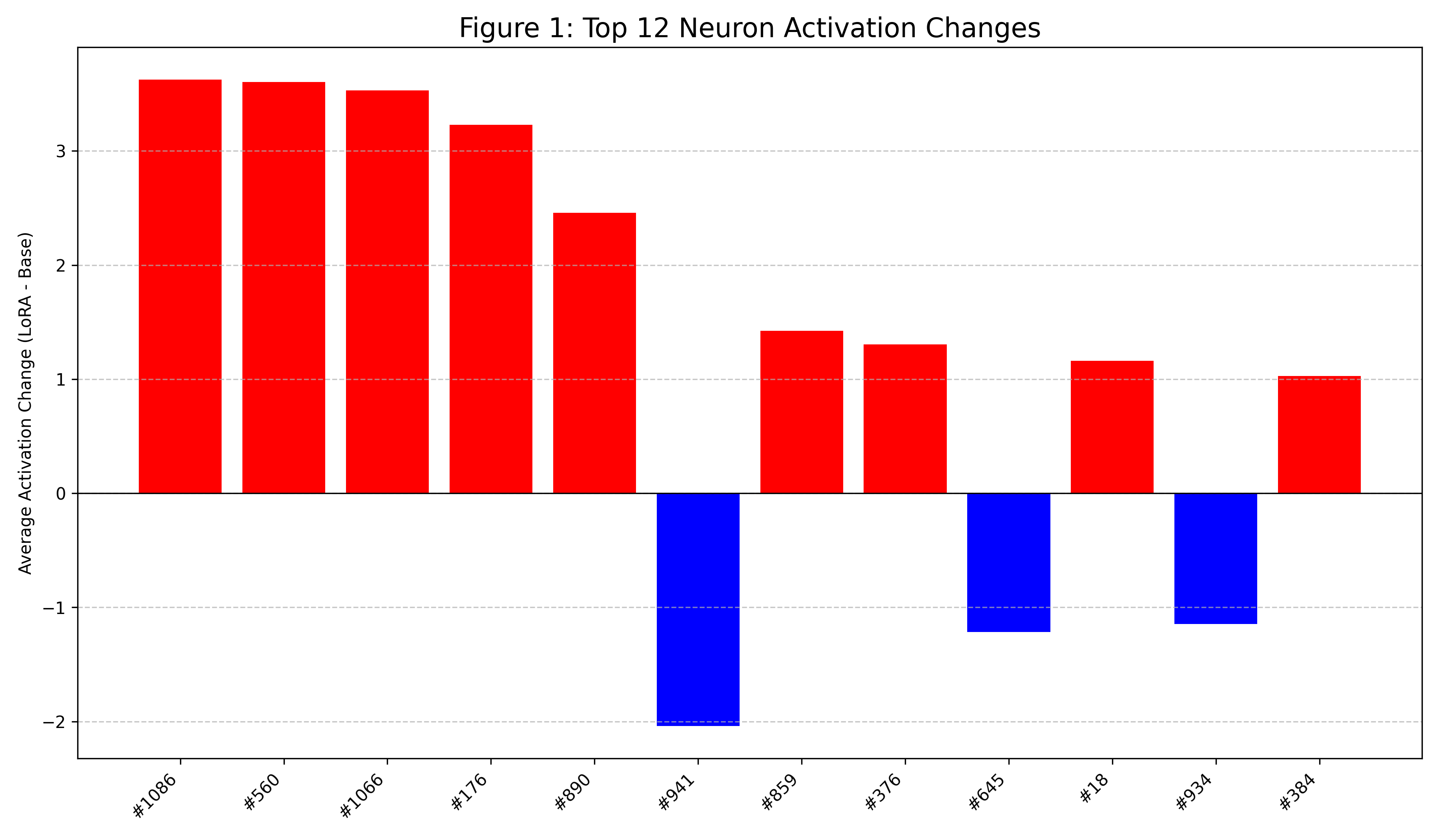}
    \caption{Average activation change for the top 12 most affected neurons after LoRA fine-tuning. Positive bars indicate amplification (increased activation), while negative bars indicate suppression (decreased activation).}
    \label{fig:neuron_changes}
\end{figure}

\begin{figure}[tbp]
    \centering
    \includegraphics[width=\textwidth]{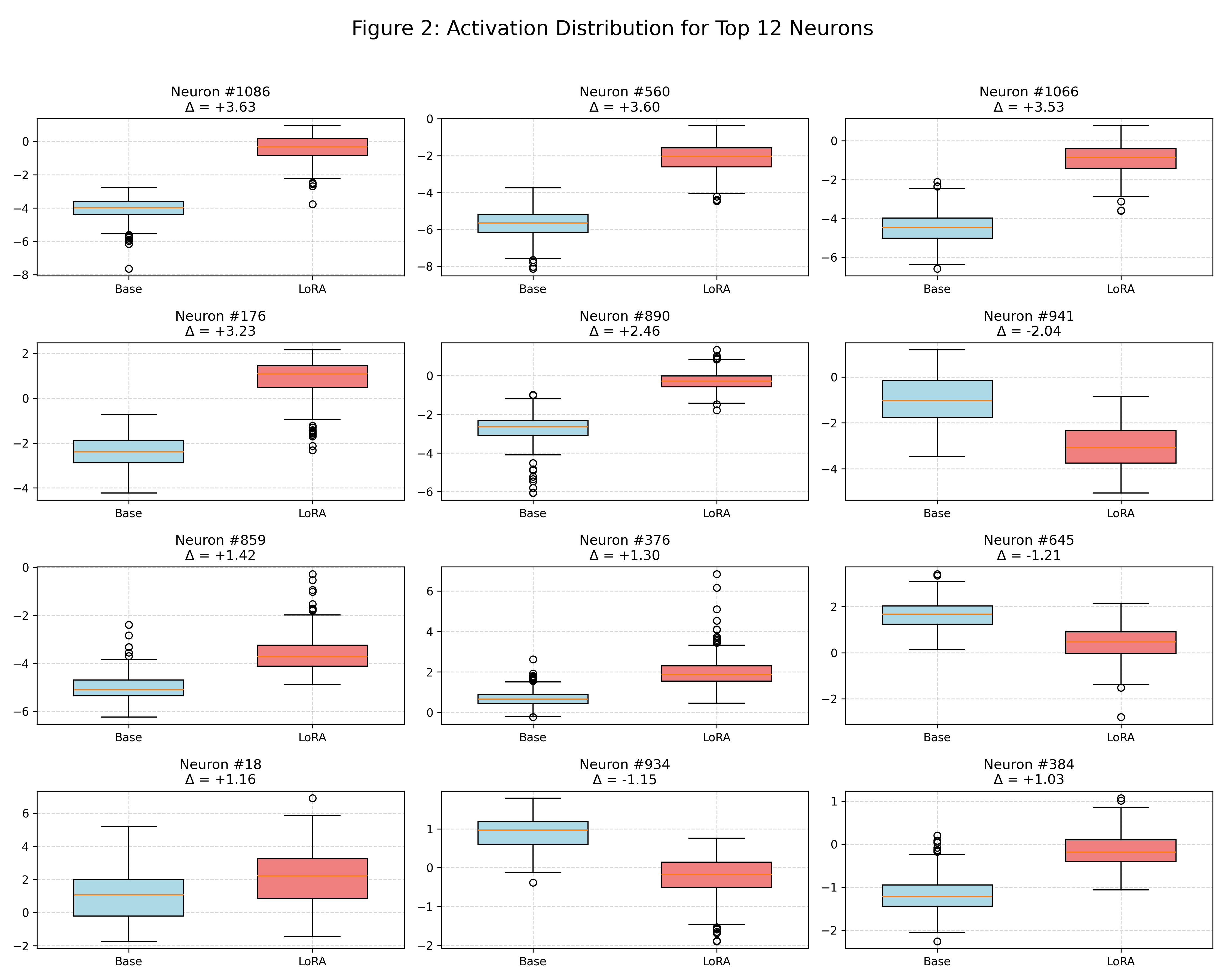}
    \caption{Boxplot comparison of activation values for the top 12 neurons, illustrating the distributional shift between the base and LoRA models.}
    \label{fig:boxplots}
\end{figure}

Our initial analysis focused on identifying neurons affected by the LoRA fine-tuning process. Figure \ref{fig:neuron_changes} provides a summary, showing the average change in activation ($|\Delta_{\text{act}}|$) for the 12 most affected neurons. A distinct pattern emerged: one set of neurons was consistently amplified (e.g., Neurons \#1086, \#560, \#1066, \#176, and \#890), showing a large positive change in activation, while another set was suppressed (e.g., Neuron \#941), showing a negative change. For a more detailed statistical view, Figure \ref{fig:boxplots} presents boxplots comparing the full activation distributions of these same neurons for the base and LoRA models. These plots visually summarize the median, interquartile range, and overall spread of activations for each neuron, providing the granular statistical information requested.

\subsection{Quantitative Effects of Neuron Silencing}
\begin{figure}[tbp]
    \centering
    \includegraphics[width=\textwidth]{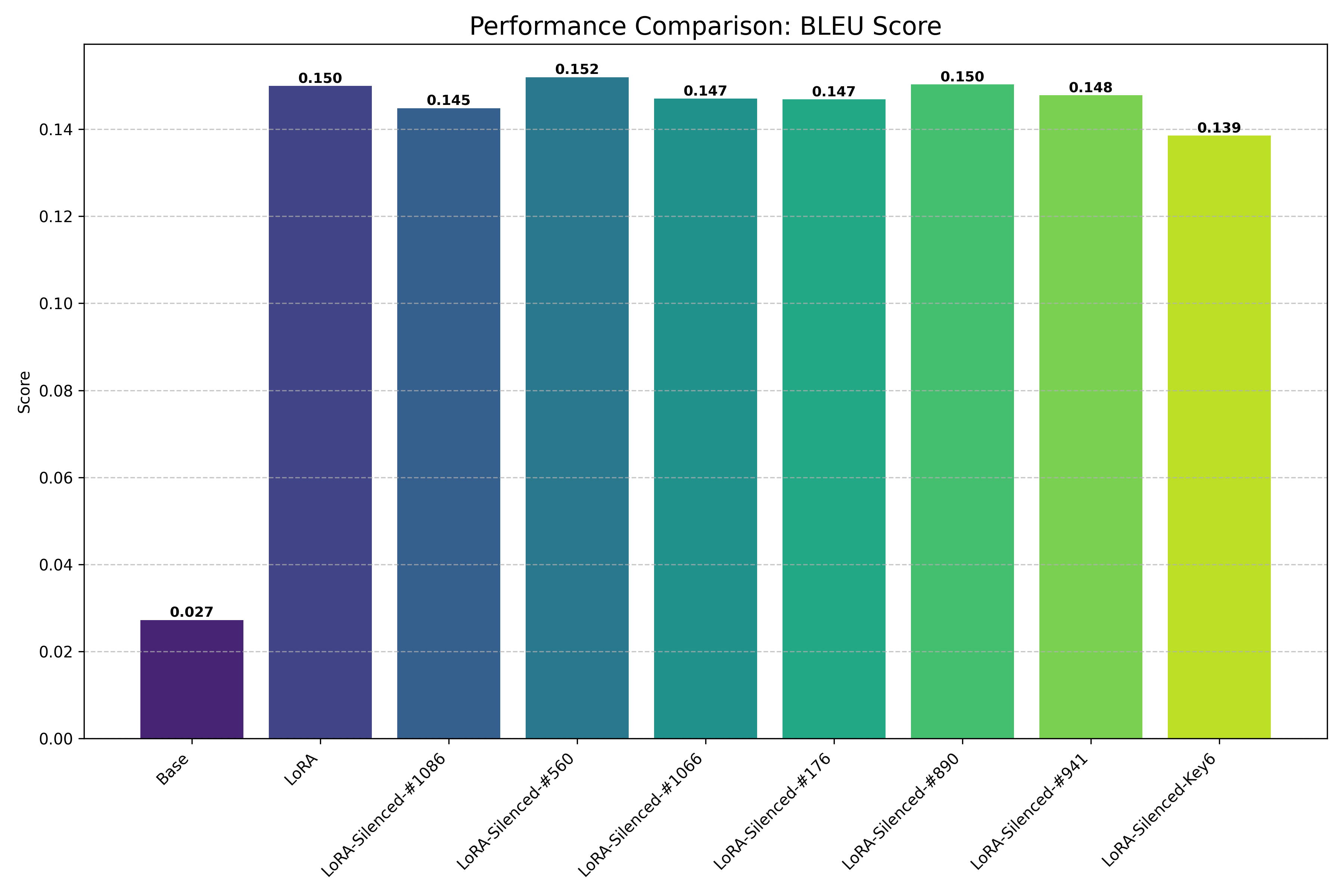}
    \vspace{0.15cm}
    \includegraphics[width=\textwidth]{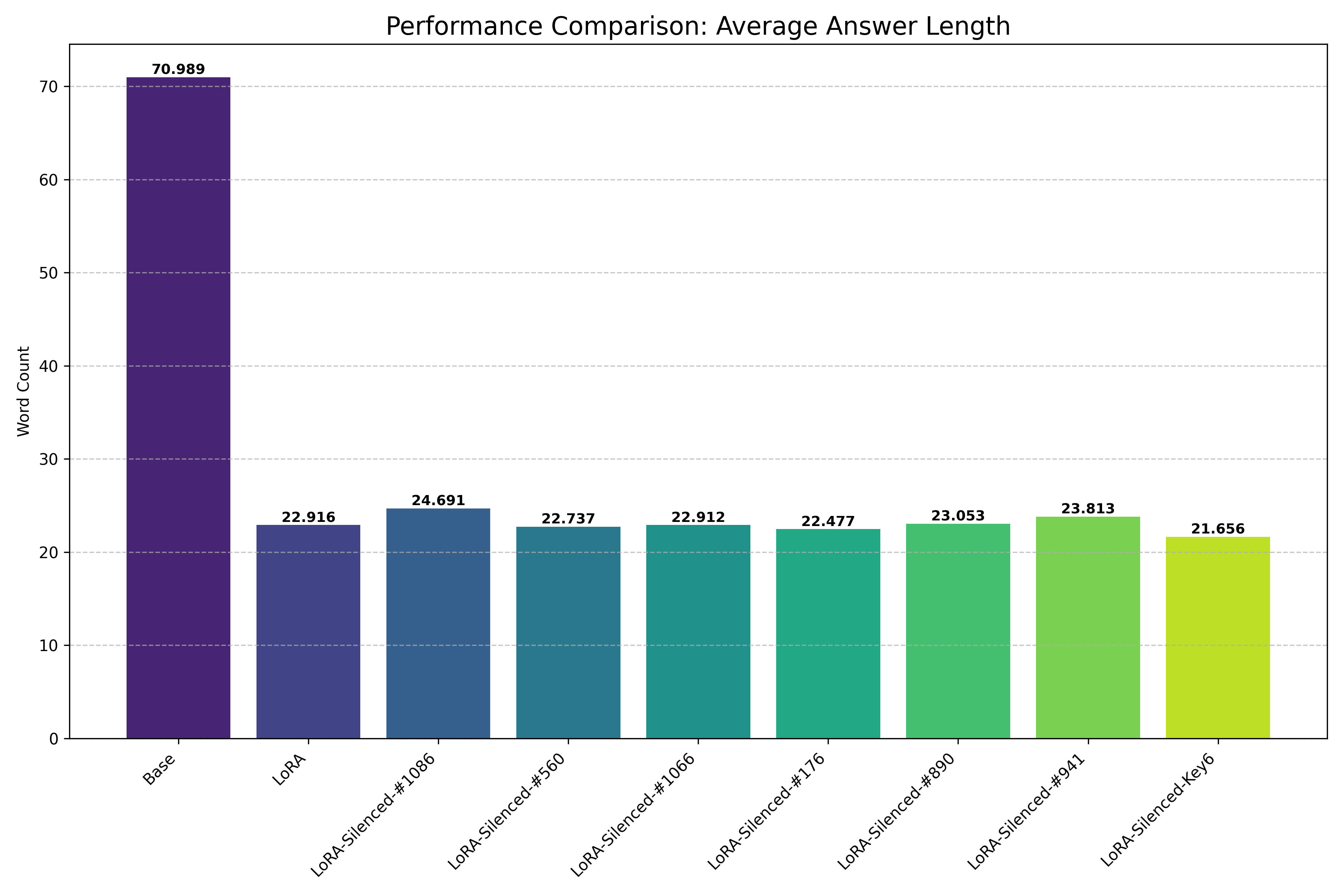}
    \caption{Comparison of model performance on the Question-Answering task: (a) BLEU Score, (b) Average Answer Length.}
    \label{fig:performance_comparison}
\end{figure}

Neuron silencing experiments were performed to measure the causal impact of the identified key neurons. Figure \ref{fig:performance_comparison}(a) presents a comparison of the BLEU scores for the base model, fine-tuned LoRA model, and several configurations in which the key neurons were individually or collectively silenced.

The LoRA fine-tuning process yielded a significant improvement over the base model, increasing the average BLEU score from 0.027 to 0.150. The effects of silencing individual neurons varied. As shown in Figure \ref{fig:performance_comparison}(a), deactivating neuron \#890 had no effect on the score (0.150), while silencing neuron \#560 led to a marginal improvement (0.152). Most other interventions caused slight performance degradation (e.g., \#1066 to 0.147). In contrast, silencing the entire group of six key neurons simultaneously (\texttt{LoRA-Silenced-Key6}) produced the largest performance decrease, reducing the average BLEU score to 0.139.

The LoRA update also induced a significant change in the response style. The base model generated verbose answers with an average length of 71.4 words, whereas the LoRA model produced more concise responses averaging 22.9 words (Figure \ref{fig:performance_comparison}(b)). Silencing interventions did not substantially alter this conciseness.

\subsection{Statistical Significance of Performance Changes}
\begin{figure}[tbp]
    \centering
    \includegraphics[width=0.8\textwidth]{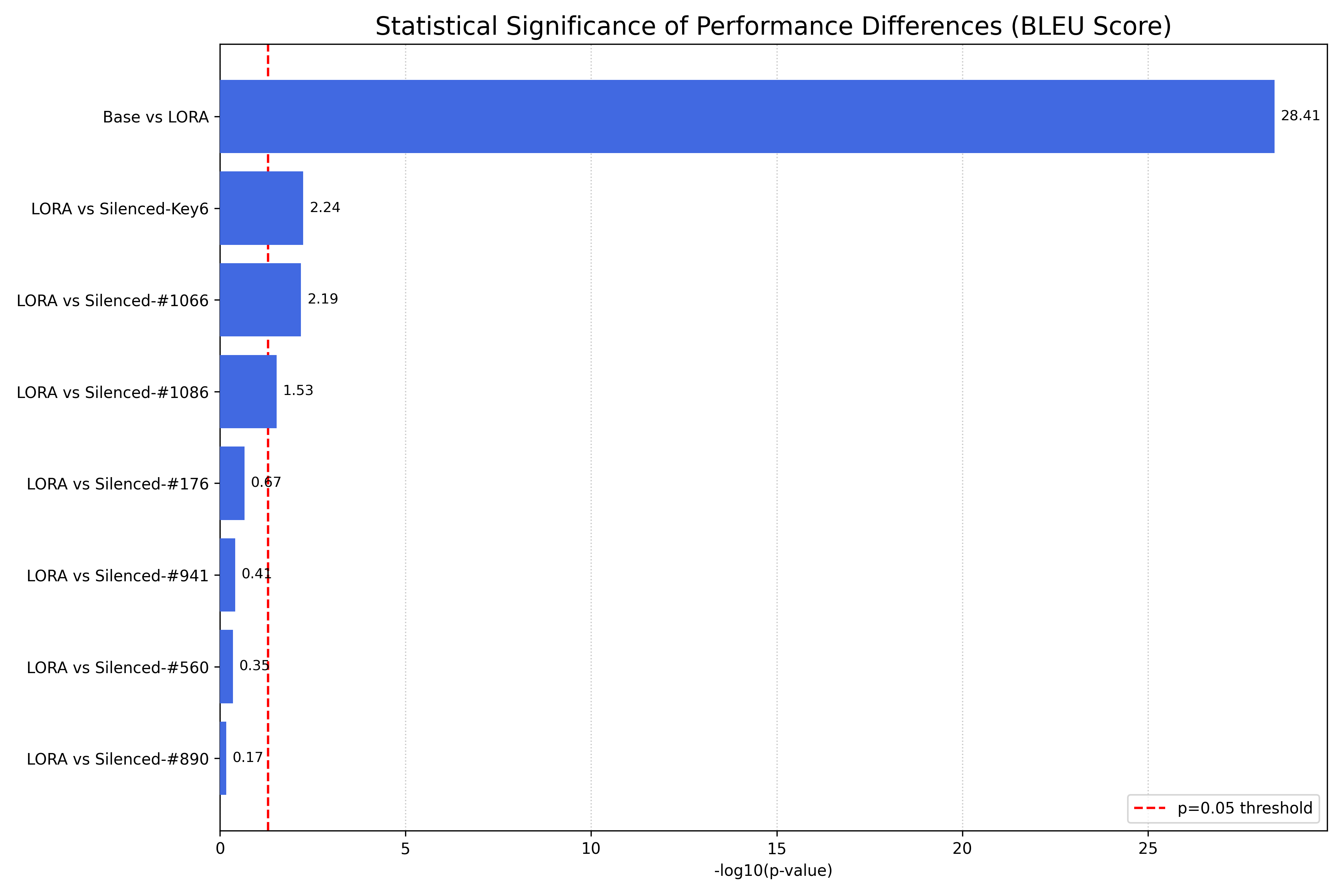}
    \caption{Statistical significance test results (-log10(p-value)). Higher bars indicate greater statistical significance. The red dashed line represents the p=0.05 threshold.}
    \label{fig:stat_tests}
\end{figure}

We performed a Wilcoxon signed-rank test to validate these observations. The results are shown in Figure \ref{fig:stat_tests}. The performance improvement from the Base to the LoRA model was highly statistically significant ($p \ll 0.05$). The performance degradation from silencing the entire key neuron group (\texttt{LoRA-Silenced-Key6}) and the most amplified individual neurons (\#1066, \#1086) were also statistically significant ($p<0.05$). The changes observed from silencing other individual key neurons did not meet the threshold for statistical significance.

\section{DISCUSSION}
\label{sec:discussion}

This section interprets the findings presented in the Results section, focusing on their scientific and practical implications for understanding AI behavior in nuclear safety contexts. First, we discuss the high-level interpretation of our findings, supporting a circuit-based view of knowledge encoding. We then performed a functional analysis of individual key neurons, attributing specific roles to the qualitative results of our silencing experiments. Finally, we directly connect these mechanistic insights to their implications for nuclear safety.

\subsection{Interpretation of Key Findings: From Independent Experts to Neural Circuits}
Our results provide evidence consistent with a circuit-based hypothesis regarding the encoding of domain expertise in a fine-tuned LLM. The finding that silencing most key neurons individually was not statistically significant while silencing the six-unit set was significant, which refutes a simple model in which each specialized neuron acts as an independent expert. Instead, it suggests that domain knowledge is a distributed property of a neural circuit, and these neurons act in concert to produce expert-level, domain-specific outputs. This circuit-based view is not just an academic distinction. It is fundamental to safety assurance in a nuclear context. Verifying the model’s reliability now means validating the integrity of distributed neural circuits responsible for safety-critical knowledge, a more complex but necessary task than simply checking individual components.

Interestingly, the most amplified neurons (\#1066, \#1086) were key exceptions, as their individual deactivation was sufficient to cause a statistically significant, albeit small, decrease in performance. This suggests a more nuanced, hybrid model in which expertise is encoded in a distributed circuit, but certain highly influential "hub neurons" may serve as critical nodes within that circuit. Nevertheless, the fact that the collective silencing had a far greater impact than silencing, even in this single most important neuron, underscores the primary importance of the circuit as a whole.

This nuanced view is further supported by intriguing anomalies observed when silencing other key neurons. For instance, deactivating neuron \#890 resulted in no performance change, while deactivating neuron \#560 paradoxically led to a slight improvement in the BLEU score. These results suggest a functional differentiation within the specialized neuron group, where individual neurons may play distinct roles such as refinement, suppression of undesirable traits, or context-specific activation, rather than uniformly contributing to knowledge generation.

\subsection{Functional Differentiation of Key Neurons}
Based on a qualitative analysis of the model's outputs when specific neurons were silenced, we can hypothesize the distinct roles of these key neurons within the larger circuit.

\begin{itemize}
    \item \textbf{Neurons \#1066 \& \#1086: Key Concept Neurons.} As the most influential "hub neurons", their function appears to be encoding and retrieval of core, domain-specific technical concepts and entities. When neuron \#1066 was silenced, the model's answer to a question about Swedish SAMG updates lost specific institutional names like "Swedish Defense Equipment (SDEk)" and defaulted to a generic, uninformative phrase like "a new version of the SAMG system". This indicates that these neurons are critical in providing the factual terminological backbone of an expert's answer. In a control room, this failure mode, losing specific terminology, could lead to dangerous ambiguity, such as referring to ‘the pump’ instead of the ‘High-Pressure Coolant Injection system’ during a critical procedure.

    \item \textbf{Neuron \#941 (Suppressed): The Generalist Language Suppressor.} This neuron's activation was decreased during LoRA fine-tuning, suggesting that it performs a function useful for general language but detrimental to technical accuracy. The Base model often produced conversational, non-committal answers "You've hit on a really important and complex question!..." and hallucinated incorrect acronyms "BWR (Bushel-Wall Reactor)". LoRA learned to suppress neuron \#941 to inhibit these undesirable generalist tendencies, allowing more specialized neurons to generate concise, factual responses. In practice, suppressing this neuron may reduce conversational filler and incorrect acronyms, thereby improving technical accuracy and conciseness.

    \item \textbf{Neuron \#176: Procedural Logic Neuron.} This neuron seems to specialize in encoding the logical relationship between different procedures or states. When asked about the transition from EOP to SAMG, silencing \#176 caused the model to produce a nonsensical answer about an ""S" sequence". However, the functional LoRA model correctly identified the condition for the switch. This suggests that neuron \#176 is crucial for representing the "if-then" logic inherent in operational guidelines. A malfunction in this neuron represents a direct operational threat, as it could cause the model to recommend an incorrect action sequence or fail to recognize the entry conditions for a critical procedure like the transition from EOP to SAMG.

    \item \textbf{Neuron \#560: Over-aggressive Generalization Neuron.} Paradoxically, silencing this neuron \textit{improved} the BLEU score. A review of its outputs suggests that it attempts to make answers more concise but sometimes does so by omitting key details. For example, when asked about maintaining water levels for jet pumps, the standard LoRA model stated it was for operating "effectively and efficiently," whereas the \#560-silenced model more accurately stated it was for operating "effectively and to prevent damage to the pump." Silencing this neuron removed a flawed simplification, resulting in a more precise and higher-scoring answer. This highlights a subtle but important safety concern: an AI that over-simplifies for the sake of conciseness could omit critical warnings or conditions, such as the risk of pump damage.

    \item \textbf{Neuron \#890: Redundant or Context-Specific Neuron.} Silencing this neuron had no observable impact on either the BLEU score or qualitative output for any question in the evaluation set. This strongly implies that its role is either fully redundant and compensated for by other neurons in the circuit or that it is highly specialized for a narrow context that was not present in our evaluation data.
\end{itemize}

\subsection{Nuclear Safety Implications of Neuron Silencing}
The quantitative performance degradation and functional roles inferred above have direct and significant implications for nuclear safety. The failure modes observed when silencing these key neurons mirror the types of errors that could lead to serious operational incidents. A qualitative analysis of the model outputs revealed three primary failure modes with significant safety consequences.

\begin{itemize}
    \item \textbf{(1) Emergency Response Procedure Failures.} As illustrated by silencing neuron \#176, disruption of procedural logic can produce incorrect EOP to SAMG transition recommendations, risking mistimed or mis-sequenced actions during accident mitigation.

    \item \textbf{(2) System Identification and Terminology Loss.} Silencing neurons \#1066 and \#1086 led to loss of specific institutional names (e.g., “Swedish Defense Equipment”), and neuron \#941 produced incorrect acronyms (e.g., “Bushel-Wall Reactor”), creating communication hazards in control room operations.

    \item \textbf{(3) Equipment Protection Logic Degradation.} In the case of neuron \#560, over-simplification omitted pump-damage-prevention conditions, posing subtle but important risks where essential cautions may be excluded.

    These failure patterns are detailed in Table \ref{tab:qualitative_examples}.
\end{itemize}

\begin{table}[tbp]
\centering
\caption{Comparison of Answer Quality After Silencing Key Neurons}
\label{tab:qualitative_examples}
\begin{tabular}{|p{0.2\textwidth}|p{0.35\textwidth}|p{0.35\textwidth}|}
\hline
\textbf{Question} & \textbf{Answer from LoRA model (Good)} & \textbf{Answer from LoRA-Silenced-Key6 model (Degraded)} \\
\hline
What is the primary function of the BWR-UH-HPCI system? & The primary function of the BWR-UH-HPCI system is to automatically provide high-pressure makeup water to the reactor vessel to maintain core cooling in the event of a small break Loss of Coolant Accident (LOCA) that does not result in rapid reactor depressurization. & The BWR-UH-HPCI system is designed to maintain the water level in the reactor pressure vessel (RPV). \\
\hline
What happens during a station blackout (SBO) event? & During an SBO, all offsite and onsite AC power is lost. The High-Pressure Coolant Injection (HPCI) and Reactor Core Isolation Cooling (RCIC) systems, which are steam-driven, become the primary means of maintaining reactor water level. DC power from station batteries is crucial for their control and operation. & A station blackout is a loss of power. The reactor will shut down. \\
\hline
Describe the function of the Automatic Depressurization System (ADS). & The Automatic Depressurization System (ADS) function is to rapidly reduce reactor pressure during a LOCA, allowing low-pressure core cooling systems like LPCI and LPCS to inject water. It is automatically initiated by signals indicating low reactor water level and high drywell pressure. & The ADS reduces pressure in the reactor. \\
\hline
\end{tabular}
\par
\vspace{1mm}
\footnotesize
\textit{Note:} HPCI, High-Pressure Coolant Injection; LOCA, Loss of Coolant Accident; RPV, Reactor Pressure Vessel; SBO, Station Blackout; RCIC, Reactor Core Isolation Cooling; DC, Direct Current; ADS, Automatic Depressurization System; LPCI, Low-Pressure Coolant Injection; LPCS, Low-Pressure Core Spray.
\end{table}

As Table \ref{tab:qualitative_examples} illustrates, the identified neural circuit encodes not only isolated facts, but also the structured, safety-oriented logic that is essential for nuclear operations. The degradation of this circuit may increase operational risk, highlighting the necessity of verifying the integrity of such circuits before deployment.

Furthermore, the phenomenon illustrated in Figure \ref{fig:performance_comparison}, where the average answer length decreased as the BLEU score increased, is consistent with improved clarity and terminological fidelity in the LoRA-adapted model relative to the base model. Although BLEU does not measure factual accuracy, the joint trend suggests more concise and precise language use. Two concrete, safety-relevant implementation scenarios are suggested by these findings. Decision support during plant transients: Recommendations could be accompanied by an interpretation dashboard indicating which key-neuron circuits are active for the current prompt, potentially improving operator trust and situational awareness. Operator training in a simulator: Circuit-activation signals could be used to compare a trainee’s response with the model’s reference pattern and to deliver targeted, context-aware feedback beyond simple correctness.

\section{Conclusion}
\label{sec:conclusion}

This research addressed the critical challenge of ensuring the reliability of LLMs in safety-critical domains like nuclear engineering. We demonstrated a replicable methodology to move beyond treating LLMs as opaque black boxes towards a mechanistic understanding of their function. Specifically, by adapting the Gemma model to the BWR domain using LoRA, we identified a sparse set of specialized neurons. Our causal interventions provided strong evidence for a circuit-based explanation by revealing the complex interplay between differentiated neuron roles and their collective impact. While silencing influential "hub" neurons like \#1066 and \#1086 individually caused a statistically significant, albeit small, degradation, this effect was far surpassed by the much larger performance drop from deactivating the entire group collectively. This demonstrates that the model's full expertise is not merely a sum of independent parts but an emergent property of the neural circuit operating in concert.

The core contribution of this study is a pathway towards achieving nuclear-grade AI assurance. The ability to identify, monitor, and causally test knowledge-bearing neural circuits offers a practical solution to the AI V\&V challenge, which has limited the deployment of advanced AI in safety-critical nuclear applications. This approach has important implications for the nuclear regulatory framework. In our case study, analysis of individual neuron functions indicates that specific failure patterns can be traced to identifiable neural components: key concept encoding neurons (\#1066, \#1086), procedural logic representation neuron (\#176), and protective-condition handling neuron (\#560). This mechanistic understanding could inform targeted acceptance criteria, such as verifying terminology precision for concept neurons or testing procedural transition logic for sequence-critical neurons, thereby establishing candidate safety benchmarks for AI system qualification. For instance, establishing baseline activation maps of key circuits during qualification, consistent with standards like IEEE Std 7-4.3.2 \cite{ieee_std_7_4_3_2}, could enable operators and regulators to continuously monitor the AI's internal reasoning and flag deviations and perform targeted re-validation after plant modifications. This aligns with human factors principles (NUREG-0711 \cite{nureg_0711}) by making AI recommendations more trustworthy and comprehensible.

However, this study had several limitations. Our reliance on the BLEU score, while justified for measuring output precision, does not verify factual accuracy. Therefore, a key area for future work is the development of domain-specific evaluation metrics that can automatically verify technical correctness against a trusted knowledge base. Furthermore, although we have identified causally important neurons, fully decoding the specific concept that each circuit represents requires more advanced techniques, such as feature visualization.

Future work will extend this methodology to other critical nuclear systems, including Pressurized Water Reactors (PWRs) and advanced reactor designs. To confirm the broader applicability of our findings, it is crucial to validate this approach across different model architectures, model scales, and technical documentation corpora.

The long-term vision is to translate these findings into practical, real-time monitoring protocols suitable for integration with plant digital infrastructure, ensuring that AI serves as a verifiable and trustworthy tool for the future of nuclear energy.

\section*{Code Availability}
The source code used to generate the results and figures in this paper is publicly available on GitHub at \url{https://github.com/sixticket/llm-xai-nuclear-paper}.

% --- CODE TO GENERATE BIBLIOGRAPHY ---
% This tells LaTeX what citation style to use. 'ans_js' is the custom ANS journal submission template bibliography style.
\bibliographystyle{ans_js}
% This tells LaTeX to read the 'bibliography.bib' file to generate the list.
\bibliography{bibliography}

\end{document}